\documentclass{article}

\usepackage{microtype}
\usepackage{graphicx}
\usepackage{subfigure}
\usepackage{booktabs} 
\usepackage{makecell}
\usepackage{multirow}
\usepackage{hyperref}

\usepackage{enumitem}
\usepackage[accepted]{icml2024}

\usepackage{amsmath}
\usepackage{amssymb}
\usepackage{mathtools}
\usepackage{amsthm}
\usepackage{xspace}

\usepackage[capitalize,noabbrev]{cleveref}

\theoremstyle{plain}

\theoremstyle{definition}

\theoremstyle{remark}

\usepackage[textsize=tiny]{todonotes}

\icmltitlerunning{Exploring Perceptual Limitations of Multimodal Large Language Models}

\begin{document}

\twocolumn[
\icmltitle{Exploring Perceptual Limitation of Multimodal Large Language Models}



\icmlsetsymbol{equal}{*}

\begin{icmlauthorlist}
\icmlauthor{Jiarui Zhang}{equal,usc}
\icmlauthor{Jinyi Hu}{equal,tsinghua}
\icmlauthor{Mahyar Khayatkhoei}{usc}
\icmlauthor{Filip Ilievski}{uv}
\icmlauthor{Maosong Sun}{tsinghua}
\end{icmlauthorlist}

\icmlaffiliation{usc}{University of Southern California, Los Angeles, California, USA}
\icmlaffiliation{tsinghua}{Tsinghua University, Beijing, China}
\icmlaffiliation{uv}{Vrije Universiteit Amsterdam, Amsterdam, Netherlands}

\icmlcorrespondingauthor{Jiarui Zhang}{jzhang37@usc.edu}

\icmlkeywords{Machine Learning, ICML}

\vskip 0.3in
]



\printAffiliationsAndNotice{\icmlEqualContribution} 

\definecolor{darkergreen}{RGB}{50,160,50}

\newcommand{\jiarui}[1]{{\color{darkergreen}(JR: #1)}}
\newcommand{\filip}[1]{{\color{orange}(Filip: #1)}}
\newcommand{\mahyar}[1]{{\color{blue}(Mahyar: #1)}}
\newcommand{\jinyi}[1]{{\color{cyan}(Jinyi: #1)}}

\newcommand{\eg}{e.g.,\xspace}

\begin{abstract}

Multimodal Large Language Models (MLLMs) have recently shown remarkable perceptual capability in answering visual questions, however, little is known about the limits of their perception. In particular, while prior works have provided anecdotal evidence of MLLMs' sensitivity to object size, this phenomenon and its underlying causes have not been explored comprehensively. In this work, we quantitatively study the perception of small visual objects in several state-of-the-art MLLMs and reveal a pervasive limitation in answering questions about small objects in images. Next, we identify four independent factors that can contribute to this limitation---object quality, size, distractors, and location---and conduct controlled intervention studies to measure the effect of each factor on MLLMs' perception. In particular, we find that lower object quality and smaller object size can both independently reduce MLLMs' ability to answer visual questions. More surprisingly, we find that the location of the object in the image and the presence of visual distractors can also significantly reduce MLLMs' question answering accuracy. Our study provides a better understanding of the perceptual limitation of MLLMs and contributes new evaluation protocols for analyzing the perception of future MLLMs. To facilitate further investigations, we release our code and data
\href{https://github.com/saccharomycetes/mllm-perceptual-limitation}{here}.

\end{abstract}

\section{Introduction}
\begin{figure}[t]
  \centering
  \includegraphics[trim=0 0 0 0, clip, width=0.47\textwidth]{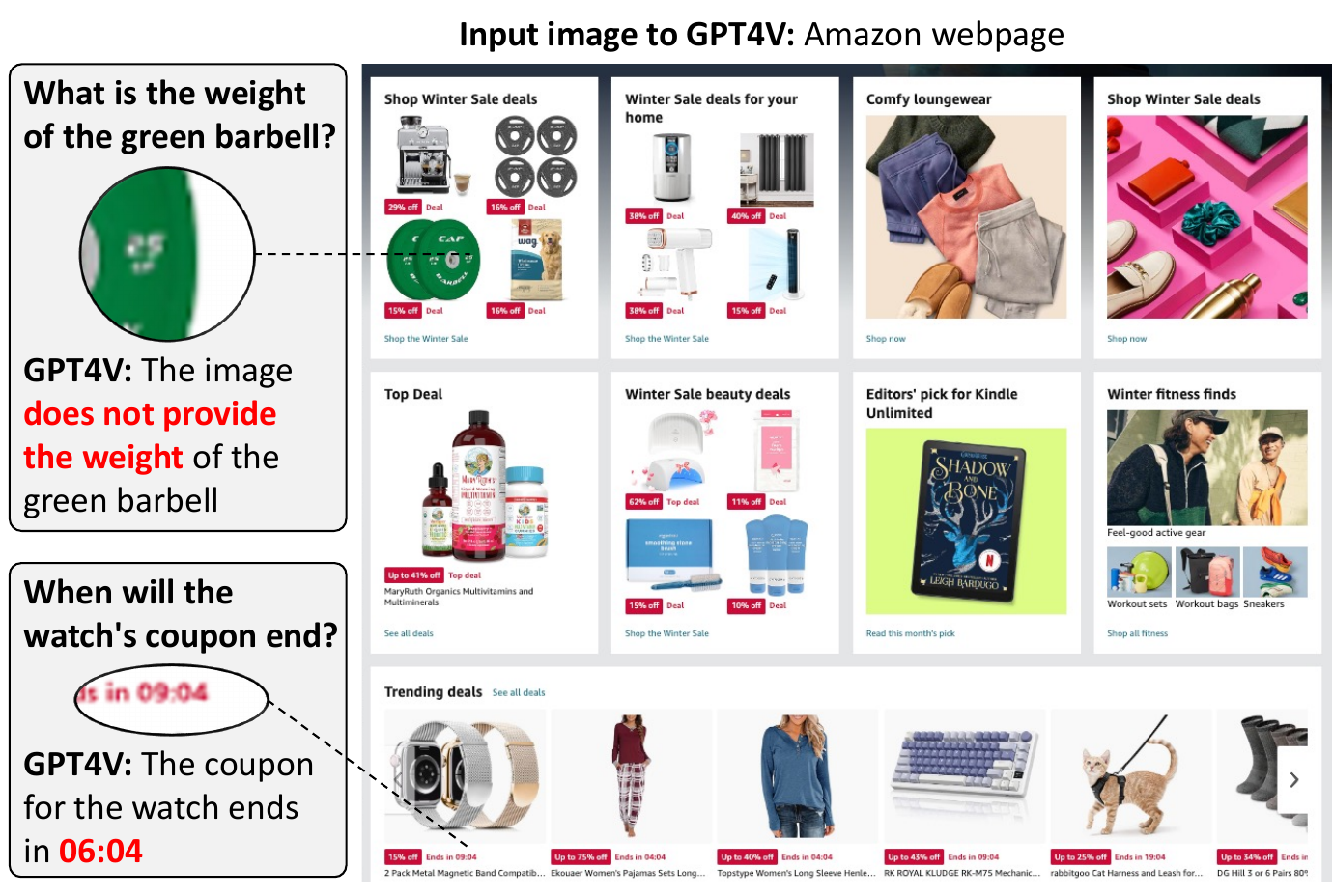}
  \caption{
  Failure cases of GPT-4V~\cite{OpenAI2023GPT4TR} in perceiving small objects when serving as web agents. Our research studies this perceptual limitation in several recent MLLMs.
  }
  \label{fig:case}
  \vspace{-0.2cm}
\end{figure}

The development of Multimodal Large Language Models (MLLMs)~\citep{OpenAI2023GPT4TR, gemini, llava, instructblip} has significantly broadened the capabilities of Large Language Models (LLMs)~\citep{OpenAI2023GPT4TR, llama}, enabling them to navigate and interpret the visual domain. Leveraging pre-trained visual encoders like CLIP-ViT~\citep{vit, clip}, MLLMs have extended the powerful textual understanding of LLMs to multimodal scenarios, such as visual question answering~\citep{blip2}, visual conversations~\citep{llava}, non-verbal reasoning~\citep{nonverbalreasoning}, and multimodal in-context learning~\citep{flamingo, mmicl}. To serve as multimodal agents~\citep{appagent, cogagent} and accomplish complex embodied tasks~\citep{palm-e, embodiedgpt}, MLLMs need to recognize and interpret visual information with different quality, size, and location, including large central objects and small peripheral pieces of text.

Despite the remarkable advancements of current MLLMs, accurately identifying small objects within images seems to remain a challenge. As~\autoref{fig:case} shows, the state-of-the-art GPT-4V~\citep{OpenAI2023GPT4TR} struggles to discern specific details like small textual descriptions. Prior research suggests that increasing the resolution of input images can generally enhance the response accuracy towards the question~\citep{qwen-vl, muffin}. Furthermore, ViCrop~\citep{zhang2023visual-crop} and V*\citep{v-star} have respectively introduced methods for image cropping and visual searching to aid MLLMs in recognizing finer details. However, the extent of this limitation and the underlying factors that lead to this challenge have not been systematically examined yet.

To bridge this gap, we quantitatively study MLLMs' perceptual sensitivity to relative object sizes and identify various visual factors that contribute to this sensitivity. We conduct a comprehensive experiment with seven state-of-the-art MLLMs on two common visual question-answering datasets, GQA~\citep{gqa} and TextVQA~\citep{textvqa}. By grouping the answers based on the relative size of target objects, we observe a significant performance drop with a decrease in object sizes, a trend that persists in all MLLMs.
Next, to identify the individual contribution of various visual factors on the MLLMs' ability to perceive small objects in images, we study four factors: \textbf{object quality}, \textbf{object size}, \textbf{object distractors}, and \textbf{object location}. Our controlled experiments yield the following findings regarding MLLMs' visual question answering performance:
\vspace{-0.15cm}
\begin{itemize}[leftmargin=*,itemsep=1pt,topsep=0pt,parsep=0pt]
    \item Object quality (sampling rate) higher than a certain threshold does not affect MLLMs' performance, and this threshold seems to align well with human perception.
    \item Smaller object size, while controlling for object quality, results in a significant decline in MLLMs' performance. This trend is less apparent in models enhanced by training on data containing annotation of smaller objects.
    \item The presence of visual distractors can reduce MLLMs' performance.
    \item The performance of MLLMs fluctuates significantly with the location of the object (visual target of the question) in the image.
\end{itemize}
\vspace{-0.15cm}
The significance of these findings is threefold.
First, our results suggest that MLLMs should be used with caution, especially when the task relies on accurately identifying visual details. Second, our findings provide novel insights for developing more reliable MLLMs, especially when dealing with data of lower quality, objects of smaller size, various distractors, and specific object positions. Third, we provide a new evaluation protocol for studying future MLLMs. This protocol can be applied, for example, to measure the robustness of an MLLM in response to different positions by showing the difference between maximum and minimum performance across different object locations.

\vspace{-0.1cm}
\section{Related work}

\textbf{Multimodal Large Language Model.}
MLLMs like GPT-4V~\citep{OpenAI2023GPT4TR} and Gemini-pro-vision~\citep{gemini} demonstrate a strong capability for visual understanding. MLLMs typically have three primary components: a vision encoder, a bridge module, and an LLM backbone \citep{muffin}. (1) \textbf{Vision Encoder}: Commonly, MLLMs utilize CLIP-ViT~\citep{clip} as the vision encoder, which divides the input image into patches and feeds them into Transformer blocks sequentially in a raster-scan order. 
(2) \textbf{Bridge Module}: The resulting visual features from the vision encoder are then either linearly projected~\citep{llava} or condensed into a fixed-sized representation~\citep{blip2} to align with the textual representation space. (3) \textbf{LLM Backbone}: The transformed visual features are then prepended to the text embedding within the LLM. 
We consider seven state-of-the-art MLLMs in this work, whose architectures are summarized in~\autoref{tab:structure}. Both BLIP-2~\citep{blip2} and InstructBLIP~\citep{instructblip} utilize the Q-Former as a bridge module, while InstructBLIP integrates instructions into the Q-Former for an instruction-awarding visual feature. LLaVA-1.5 \citep{videollava} projects the visual feature from ViT into the LLM space with an MLP layer. Qwen-VL-Chat~\citep{qwen-vl} chooses a larger vision encoder ViT-bigG and a one-layer cross-attention module to perceive visual features. Fuyu-8B~\citep{fuyu-8b} uniquely removes the external vision encoder, directly incorporating pixel information into the language decoder. 
The training of MLLMs typically undergoes an initial pre-training on extensive image-text datasets such as LAION~\citep{laion}, followed by specialized multimodal instruction tuning~\citep{llava}. 
Enhancements in MLLMs have been pursued through various means, including increasing image resolution~\citep{muffin}, scaling data and model size~\citep{cogvlm}, extending to multilingual context~\citep{viscpm}, and introducing interleaved data formats~\citep{vila}.

\begin{table}[t]
\centering
\caption{Architectural overview of the MLLMs used in this study.}
\scalebox{0.85}{
\begin{tabular}{l|ccc}
\toprule
Model         & \makecell[c]{Vision \\ Encoder} & \makecell[c]{Bridge \\ Module}    &  \makecell[c]{LLM \\ Backbone}    \\ \midrule
BLIP-2        &  ViT-g                          &  Q-Former                         &     Flan-T5$_{\mathrm{XXL}}$      \\ 
InstructBLIP  &  ViT-g                          &  Q-Former                         &     Vicuna-13B                    \\ 
LLaVA-1.5     &  ViT-l                          &  Linear                           &     Vicuna-13B                    \\ 
Qwen-VL-Chat  &  ViT-bigG                       &  Resampler                        &     Qwen-7B                       \\ 
Fuyu-8B       &  -                              &  -                                &     Persimmon-8B                  \\ \midrule
GPT-4V &  \multicolumn{3}{c}{Not released}          \\
Gemini-Pro-Vision &  \multicolumn{3}{c}{Not released}          \\
\bottomrule
\end{tabular}}
\label{tab:structure}
\end{table}

\textbf{Robustness Analysis to MLLMs.} 
The capabilities of MLLMs have been evaluated using general benchmarks like the traditional VQA benchmark VQAv2~\citep{VQA} and GQA~\citep{gqa}, alongside newer benchmarks such as MM-Bench~\citep{mmbench} and MMMU~\citep{mmmu}. Some works have shown that MLLMs suffer from object hallucination~\citep{pope, rlhf-v} and a lack of robustness in processing visual details~\citep{zhang2023visual-crop}. The MMVP benchmark~\citep{eyes} further highlights these visual shortcomings, particularly emphasizing the discrepancy between the embedding spaces of CLIP and the vision-only self-supervised space of DINOv2. The V* algorithm~\citep{v-star} offers an innovative approach with its LLM-guided visual search method, specifically targeting the focus on visual details. Our paper builds upon these insights, quantitatively exploring MLLMs' performance in handling visual details, considering four distinct factors.

\section{Can MLLMs Perceive Small Objects?}
\label{sec:motivation}

\begin{figure}[t]
  \centering
  \includegraphics[trim=0 0 0 0, clip, width=0.45\textwidth]{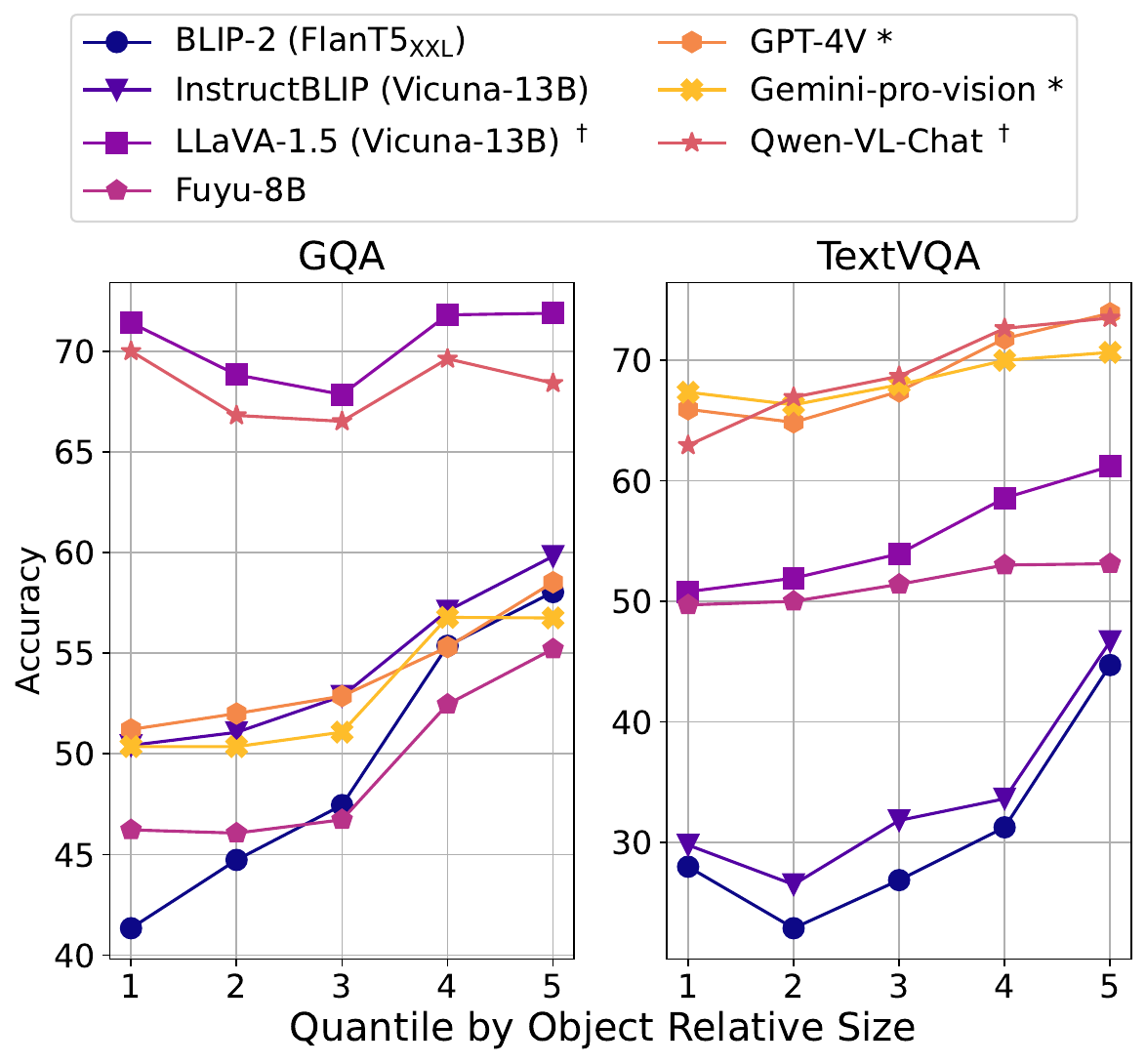}
  \caption{
  The performances of multiple popular MLLMs on GQA and TextVQA show a clear positive correlation with relative size of target objects. The accuracy is computed with \textbf{inclusion match}.
  *A small part of questions is skipped due to safety policy of API models. $^\dagger$The model has been reported to be trained on the dataset.
  }
  \label{fig:motivation}
  \vspace{-0.2cm}
\end{figure}

\begin{table}[t]
\centering
\small
\caption{Intervals of the number of pixel values (\textbf{\# Pixels}) within the bounding box after the input images are unified to size \textbf{224$\times$224} on 5 data quantiles of GQA and TextVQA tested in~\cref{sec:motivation}, and the average number of object/textual distractors (\textbf{\# D}) in two datasets. \\}
\begin{tabular}{c|l|l|l|l}
\toprule
\textbf{Data} & \multicolumn{2}{c|}{\textbf{GQA}} & \multicolumn{2}{c}{\textbf{TextVQA}} \\
\textbf{Quantile} & \# Pixels & \# D & \# Pixels & \# D \\
\midrule
1 & [100, 1409]   & 19.2 & [26, 80]    & 14.9 \\
2 & [1409, 5043]  & 18.1 & [80, 143]   & 13.8 \\
3 & [5045, 11967] & 17.4 & [143, 296]  & 13.1 \\
4 & [11971, 24571]& 16.7 & [296, 697]  & 11.7 \\
5 & [24588, 50176]& 14.6 & [698, 50176]& 7.6 \\
\bottomrule
\end{tabular}
\label{tab:ranges}
\vspace{-0.4cm}
\end{table}

Recent anecdotal evidence~\cite{zhang2023visual-crop} suggests that MLLMs face challenges in perceiving small visual details compared to larger ones. Inspired by this research, we conduct an extensive quantitative experiment to study the sensitivity to size of recent SOTA MLLMs on two standard VQA benchmarks.
We evaluate the seven representative models shown in~\autoref{tab:structure} on two prominent visual question-answering datasets, GQA~\cite{gqa} for compositional reasoning on real-world objects and TextVQA~\cite{textvqa} for reading and comprehending texts presented in the real-world image. Both datasets offer the advantage of bounding box annotations, pinpointing areas of interest within images. 
For GQA, we aggregate bounding boxes encompassing all related objects. For TextVQA, we focus on the bounding box with the highest textual similarity to the ground-truth answer. To facilitate a nuanced assessment, we categorize these datasets into quintiles based on the relative size of the target area. The accuracy is measured via inclusion match~\citep{multimodalocr} (see~\cref{sec:appendix_motivation} for exact match results).

Our findings, depicted in~\autoref{fig:motivation}, demonstrate a consistent issue across all models: a marked decline in processing accuracy for smaller visual elements. Such a trend is most notable in BLIP-2, whose performance gap across different quantiles is 16.71\% and 21.83\% on GQA and TextVQA, respectively.
In addition, the two leading closed-source API models, GPT-4V and Gemini-provision, have a 7.32\% and 6.39\% performance gap on GQA and 9.05\% and 3.31\% on TextVQA, respectively, also exhibiting performance gaps.

To better understand the underlying reasons affecting the performance in~\autoref{fig:motivation}, we compute the range of the number of pixels within each quantile after unifying the input images to 224$\times$244 in~\autoref{tab:ranges}. For models with higher resolutions, the number of pixels increases proportionally. In TextVQA, the numbers of pixels of the target texts on the first three quantiles are notably limited, indicating that the information of the target object only accounts for a small portion of the whole input. Meanwhile, the limited number of pixels also causes a low image quality presenting the target object. Furthermore, we also compute the average number of distracting OCR tokens (OCR tokens that are not related to the answer) in each quantile of TextVQA and the average number of distractor objects in GQA, the decreased number of distractors could also potentially contribute to~\autoref{fig:motivation}.

Based on the analysis, we summarize the underlying reasons for such limitations as four potential factors. The following section first systematically formulates these factors and then explores their impact on the MLLMs' capacity to recognize and interpret small visual objects.

\begin{figure}[t]
  \centering
  \includegraphics[trim=0 0 0 0, clip, width=0.45\textwidth]{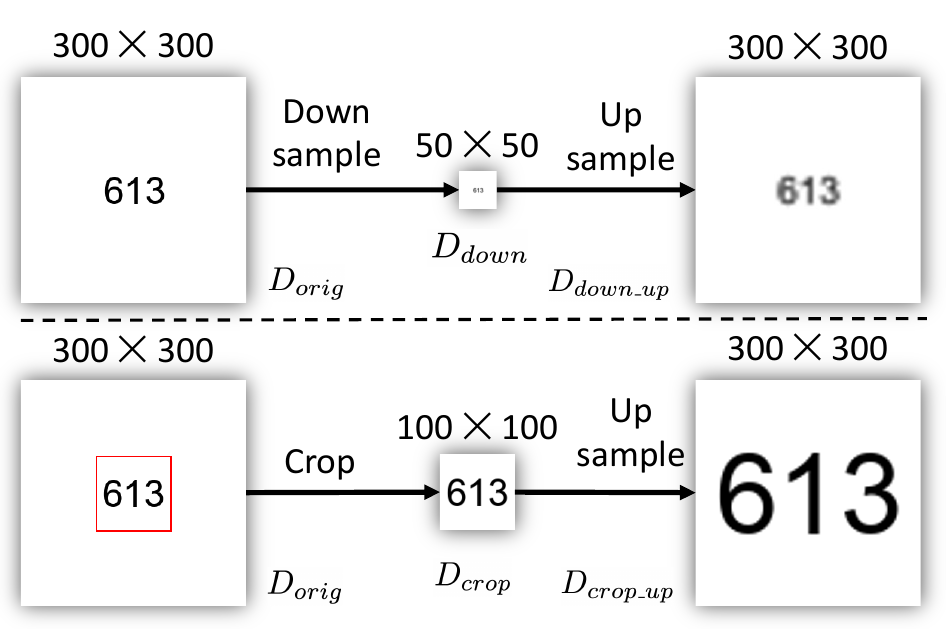}
  \caption{
  An illustration of the Downsample-Upsample (upper) and Crop-Upsample (lower) procedure described in~\cref{sec:quality} and~\ref{sec:size}. The upper process reduces object quality 6 times while keeping the same size and position. The lower increases object size three times while keeping the object quality.
  }
  \label{fig:process}
  \vspace{-0.2cm}
\end{figure}

\section{What Factors Affect MLLMs' Perception of Small Objects?}
\label{sec:problem_formulation}

We focus on the following four factors: \textit{object quality, object size, object distractors, and object location}. While the identified factors are by no means exhaustive, they aim to illuminate some of the fundamental perceptual limitations of current MLLMs, thereby informing both practical applications and future enhancements of these models.

\begin{figure*}[t!]
  \centering
  \includegraphics[trim=0 0 0 0, clip, width=0.95\textwidth]{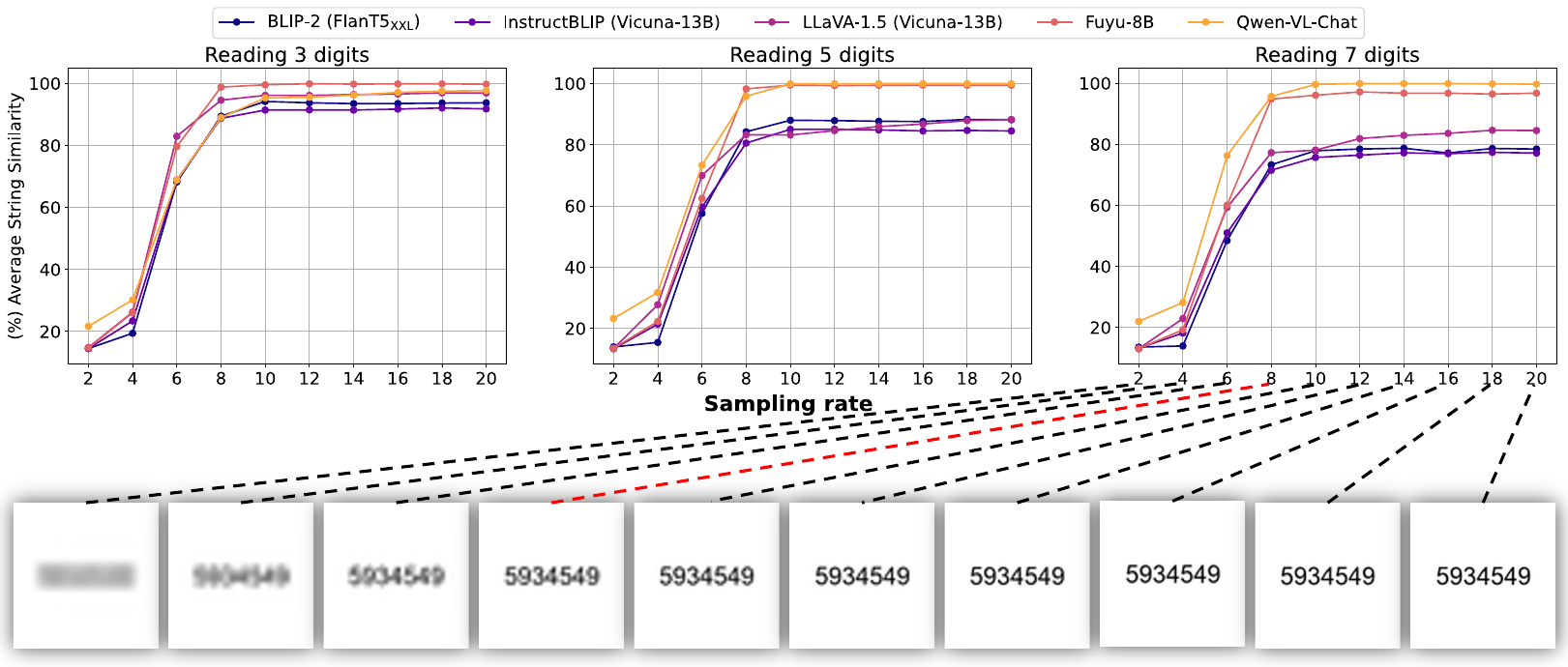}
  \caption{
  The effect of changing text sampling rate (quality) on model's performance of reading texts while keeping the size of the text. It is noticeable that from the sampling rate of 8 (marked as \textcolor{red}{red}), the image starts to become fully recognizable as `5934549'.
  }
  \label{fig:quality}
  \vspace{-0.1cm}
\end{figure*}

\textbf{Object Quality.} We define quality as the original \textbf{sampling rate of an object} (in pixels per inch, or pixels per vector graphic range), that is, the original resolution of an object in a given image. To vary object quality, we adopt a downsample-upsample strategy on an original high-resolution image of the object, which is illustrated in the upper part of~\autoref{fig:process}. Starting from an original 300-pixel by 300-pixel raster image of a vector graphic digit ($D_{orig}$), we reduce its quality six times by down-sampling that raster image to 50 pixels by 50 pixels ($D_{down}$). Then we upsample the $D_{down}$ six times, and the resulting $D_{down\_up}$ reaches the same image size with $D_{orig}$, but a six times lower sampling rate. Note that image upsampling does not inherently change the sampling rate of the object despite the increase in pixel values. In this paper, we use the terms `sampling rate' and `quality' interchangeably.

\textbf{Object Size.} The object size is defined as the number of pixels that belong to an object in the input image to MLLMs. Note that we can modify the object size while keeping its quality constant by upsampling the object to the desired size. To this end, we adopt a \textbf{crop-upsample} strategy, as is illustrated in~\autoref{fig:process} (lower). 
Given a 300-pixel by 300-pixel raster image of a digit (of a particular quality due to the original sampling rate), we crop the $D_{orig}$ at the center to 100 pixels by 100 pixels ($D_{crop}$). Then we upsample the $D_{crop}$ three times, resulting $D_{crop\_up}$ with the same sampling rate and image pixel size with $D_{orig}$, while having a three times larger object size.

\textbf{Object Distractors.} Object distractors are objects that belong to the same distribution as a target object of interest (\eg, other numbers when the object of interest is a particular number in the image).

\textbf{Object Location.} 
Current MLLMs share the same manner for image processing, where a complete image is divided into numerous patches, which are subsequently transformed into individual image tokens.
Formally, the input image \( \mathrm{x} \in \mathbb{R}^{H \times W \times C} \) with spatial dimensions \( (H, W) \) and \( C \) color channels is first reshaped into $2D$ patches \( \mathrm{x_p}\in \mathbb{R}^{N \times P^2 \times C} \), and the resulting $N$ image patches are mapped to $N$ token embedding as the input of transformer architectures. Given the architecture, an input object could be cut by image patch boundaries and divided into different image patches.
In light of this, we investigate two complementary location-related factors: the global location on the image and the local patch boundary cut on the target object.

\vspace{-0.2cm}
\subsection{Experimental Setup}
\textbf{Text-Reading Objective.}
In our experiments, we focus exclusively on the text-reading ability of MLLMs. This decision is driven by the idea that text reading involves recognizing diverse shapes and their spatial relationships, providing a clear and definitive framework for assessment. Compared to other visual tasks like identifying object colors or types, text recognition offers reduced ambiguity in evaluation. To facilitate controlled comparisons, we use synthetic digital texts, rendered in the widely used Arial sans-serif font, and overlaid on plain white backgrounds. Here, the `sampling rate' is defined in terms of the font size used during text creation, which correlates with the vertical pixel count of the text characters.
During the evaluation, the accuracy of the MLLMs’ responses is assessed against the actual text in the images using Gestalt Pattern Matching (\textbf{GPM})~\citep{gpm}. This metric is a widely used smooth metric for OCR task assessments.

\textbf{Evaluated Models.}
Due to the prohibitive cost of running granular experiments on commercial MLLMs, we will consider the five open-source models as representative examples of current MLLMs: BLIP2~\citep{blip2}, InstructBLIP~\citep{instructblip}, LLaVA-1.5~\citep{llava1.5}, Qwen-VL-Chat~\citep{qwen-vl} and Fuyu-8B~\citep{fuyu-8b}. The architectures of five models are introduced in \cref{sec:problem_formulation}. Notably, BLIP-2 has not been explicitly trained on OCR-oriented tasks, relying instead on image-text pairs with text annotations within the images. InstructBLIP and LLaVA-1.5 have undergone training on several OCR-oriented tasks, including OCR-VQA~\citep{ocrvqa} and TextCaps~\citep{textcaps}. Qwen-VL-Chat, having been trained on a substantial 25M OCR-oriented dataset, demonstrates enhanced OCR capabilities, and is thus referred to as an OCR-enhanced-MLLM in our analysis. The training specifics for Fuyu-8B are not publicly disclosed, but based on its performance, we presume its OCR training to be similar to that of Qwen-VL.

\vspace{-0.1cm}
\subsection{Quality Sensitivity Study}
\label{sec:quality}

Our goal in this section is to study the ability of MLLMs in reading small text of varying quality (sampling rates). We adopt the \textbf{Downsample-Upsample} strategy which is described in~\autoref{fig:process} and construct a dataset with a sampling rate from 2 to 20 in increments of 2, examples are shown at the bottom of~\autoref{fig:quality}. Our experimental tasks involve reading 3, 5, and 7 digits, signifying three tiers of task complexity, placed at the center of an image. Each tier includes 500 random numbers to read. We prompt MLLMs with the question \textit{``What is the number on the image?''}.

\textbf{MLLMs' response to object quality is threshold-dependent.}
 As shown in~\autoref{fig:quality}, we observed a significant improvement in the MLLMs' performance as the sampling rate increased from 4 to 8. However, after this point, the performance stabilized with increasing sampling rate, indicating a threshold-dependent trend in the MLLMs' ability to read text of varying qualities.

\textbf{The threshold is universal and aligns well with human perception.}
Remarkably, the threshold of a sampling rate of 8 is consistently observed across all MLLM models, irrespective of their text recognition capabilities and the varying levels of task complexity. This threshold seems to be consistent with human perceptual ability, as it becomes hard to read text below this threshold for our own eyes. 
These findings suggest that the MLLMs' response to image quality is more influenced by the intrinsic properties of the images rather than the internal differences among the MLLMs. Considering this threshold-dependent performance improvement, the continuous improvement in performance within image size observed in~\autoref{fig:motivation} cannot be solely attributed to image quality improvements. In the following sections, we conduct further experiments to investigate other factors that can affect the perception of small objects by MLLMs.

\subsection{Size Sensitivity Study}
\label{sec:size}
\begin{figure*}[t!]
  \centering
  \includegraphics[trim=0 0 0 0, clip, width=0.99\textwidth]{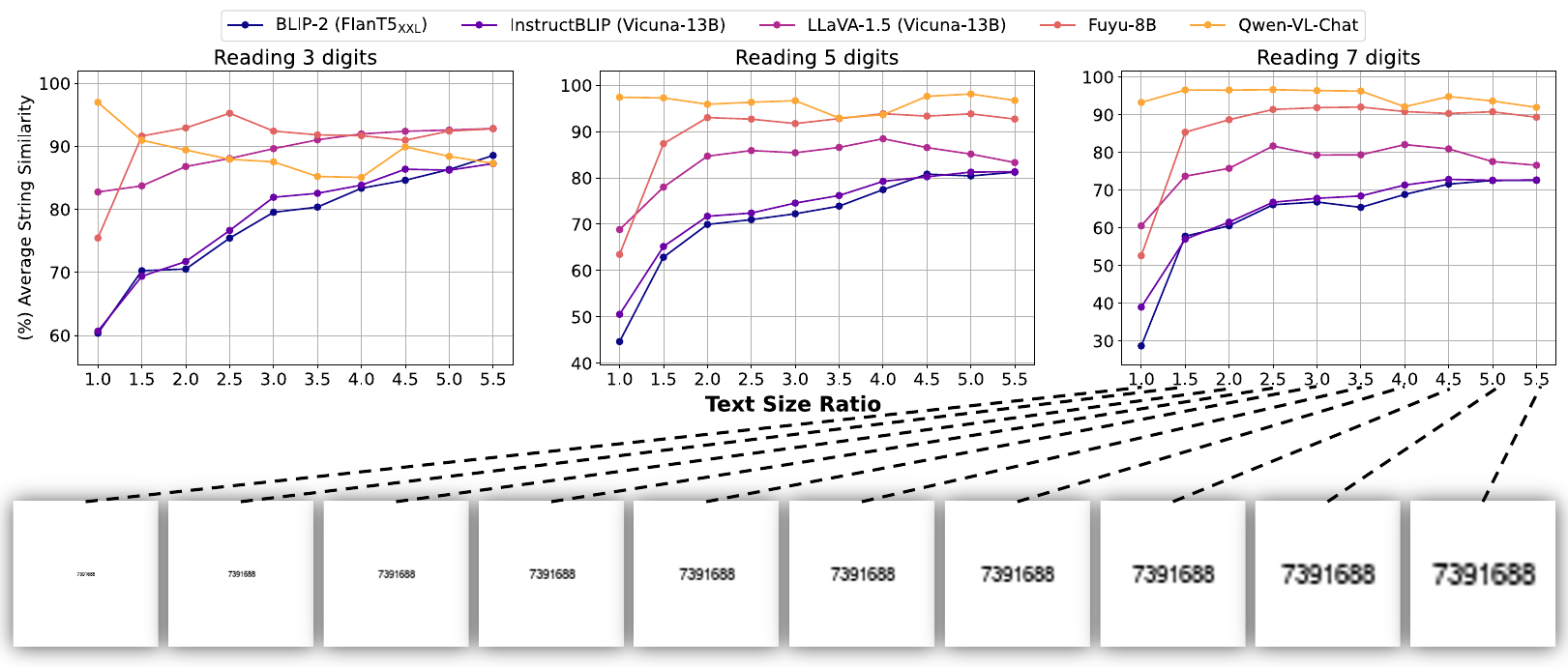}
  \caption{
  The effect of changing text size on model's performance of reading texts while keeping the sampling rate of the text. 
  }
  \label{fig:size}
  \vspace{-0.1cm}
\end{figure*}
\begin{figure}[t]
  \centering
  \includegraphics[trim=0 10 0 0, clip, width=0.45\textwidth]{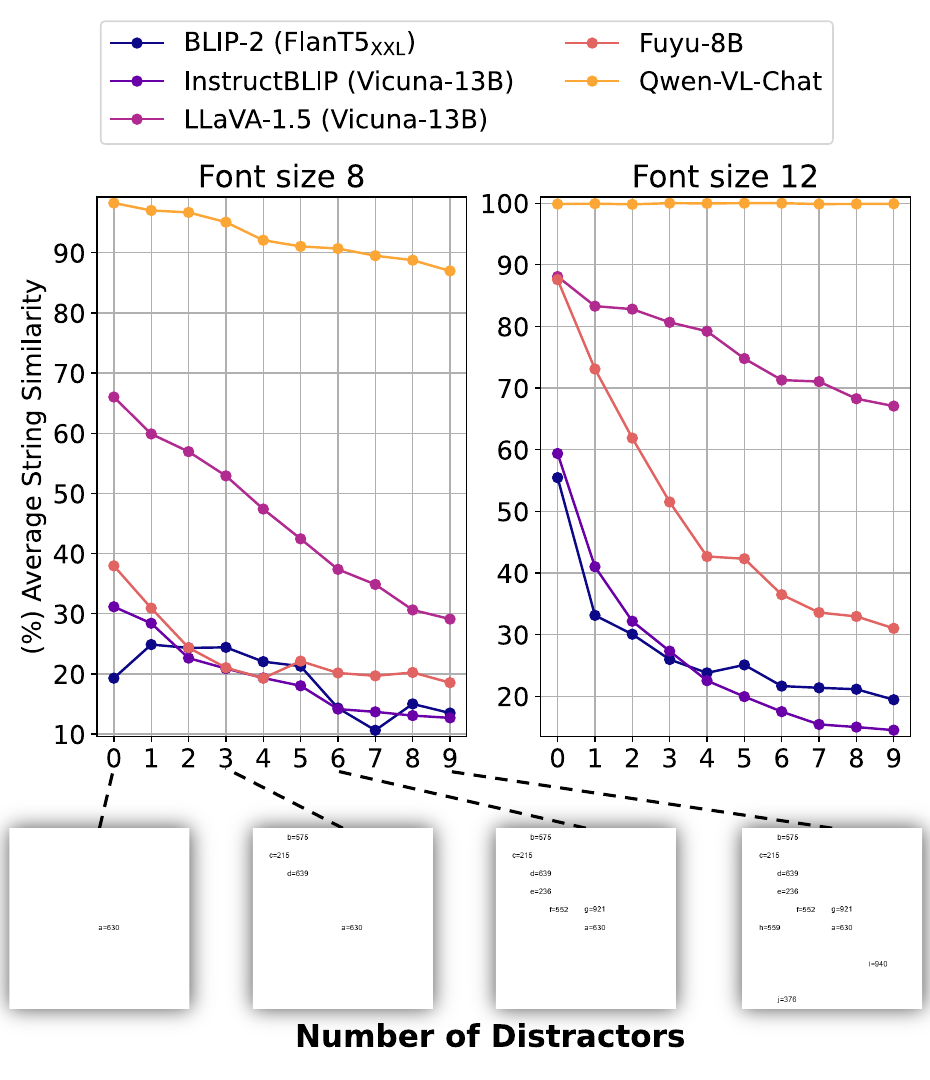}
  \caption{
  The effect of changing the number of distractors on MLLMs' performance of reading texts.
  }
  \label{fig:distract}
  \vspace{-0.3cm}
\end{figure}

In the preceding section, we observed that the sampling rate of text does not significantly challenge MLLMs after a certain threshold. This leads us to inquire about the impact of object size on MLLMs' performance with a fixed sampling rate (quality). To explore this, we follow the~\textbf{Crop-Upsample} strategy described in~\autoref{fig:process}. Specifically, for $D_{orig}$, we place an 8-font size text in the center of the image, then in $D_{crop\_up}$ the original text is enlarged 1 to 5.5 times, with a step of 0.5, illustrated at the bottom of~\autoref{fig:size}. The tasks include recognizing 500 random numbers with 3, 5, and 7 digits following~\cref{sec:quality}. We prompt MLLMs with the question \textit{``What is the number on the image?''}.

\textbf{At a fixed object quality, most MLLMs perform better at recognizing larger objects.}
As shown in~\autoref{fig:size}, except for the OCR-enhanced model Qwen-VL-Chat, the performance of MLLMs improves with the increase of object size while maintaining a constant quality (sampling rate). Notably, the performance trajectory of Fuyu-8B exhibits a significant enhancement in the early stages of size increase. In contrast, BLIP-2 and InstructBLIP show a more gradual improvement in performance with increasing object size. LLaVA-1.5, however, demonstrates a relatively stable performance across varying sizes, indicating a lesser sensitivity to changes in object size. Furthermore, we observe that for tasks with greater complexity (recognizing more digits), the increase in object size has a larger impact on the models' accuracy.
This phenomenon may be attributed to two reasons. First, larger object sizes occupy more image patches. These patches translate into transformer tokens, which, during the self-attention mechanisms of the transformer architecture, allow for a more extensive fusion of information. Second, the majority of MLLM image-text matching data for pre-training, only present textual descriptions for the main visual components in the image which are often larger, diminishing their capability of perceiving smaller objects. The second reason is supported by the fact that the OCR-enhanced model Qwen-VL-Chat, which is trained on large-scale synthetic data with 41 English fonts and 11 Chinese fonts, maintains its accuracy when processing smaller objects.

\vspace{-0.2cm}
\subsection{Distractor Sensitivity Study}
\label{sec:distractor}

\begin{table}
\centering
\small
\caption{The input image patch number and patch size of the MLLMs considered in our experiment. *Fuyu-8B has a fixed patch size of 30$\times$30 but does not have a fixed patch number. We set it to 10$\times$10 in our experiment.}
\begin{tabular}{l|lll}
\toprule
Model         & \makecell[c]{Patch Number} &   \makecell[c]{Patch Size} &   \makecell[c]{Resolution} \\ \midrule
BLIP-2        &  16$\times$16              &   14$\times$14             &   224$\times$224\\ 
InstructBLIP  &  16$\times$16              &   14$\times$14             &   224$\times$224\\ 
LLaVA-1.5     &  24$\times$24              &   14$\times$14             &   336$\times$336\\ 
Qwen-VL-Chat  &  32$\times$32              &   14$\times$14             &   448$\times$448\\ 
Fuyu-8B       &  10$\times$10*             &   30$\times$30             &   300$\times$300*\\
\bottomrule
\end{tabular}
\label{tab:patchsize}
\vspace{-0.2cm}
\end{table}

\begin{figure*}[t]
  \centering
  \includegraphics[trim=0 0 0 0, clip, width=0.99\textwidth]{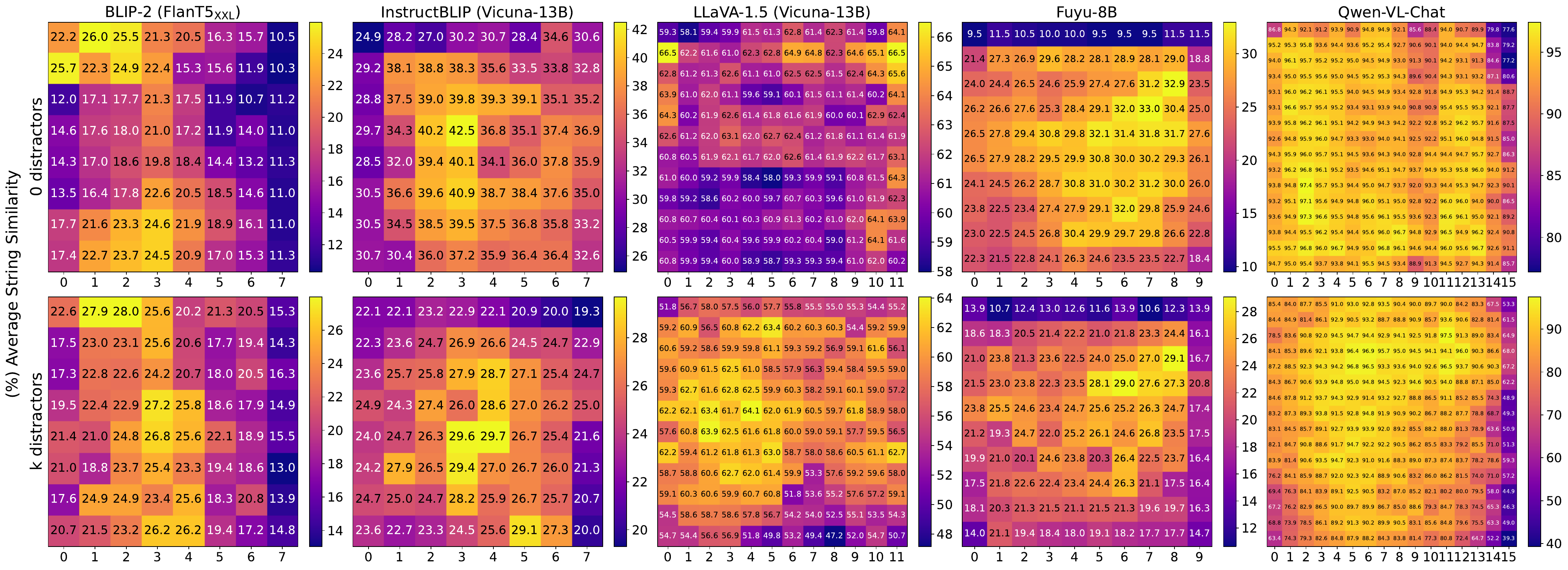}
  \caption{
  The effect text (number) location in the image on MLLMs' ability to read the text correctly, with and without distractors (bottom and top, respectively). Higher values are presented in lighter colors. 
  }
  \label{fig:position}
  \vspace{-0.1cm}
\end{figure*}
\begin{figure*}[t]
  \centering
  \includegraphics[trim=0 0 0 0, clip, width=0.91\textwidth]{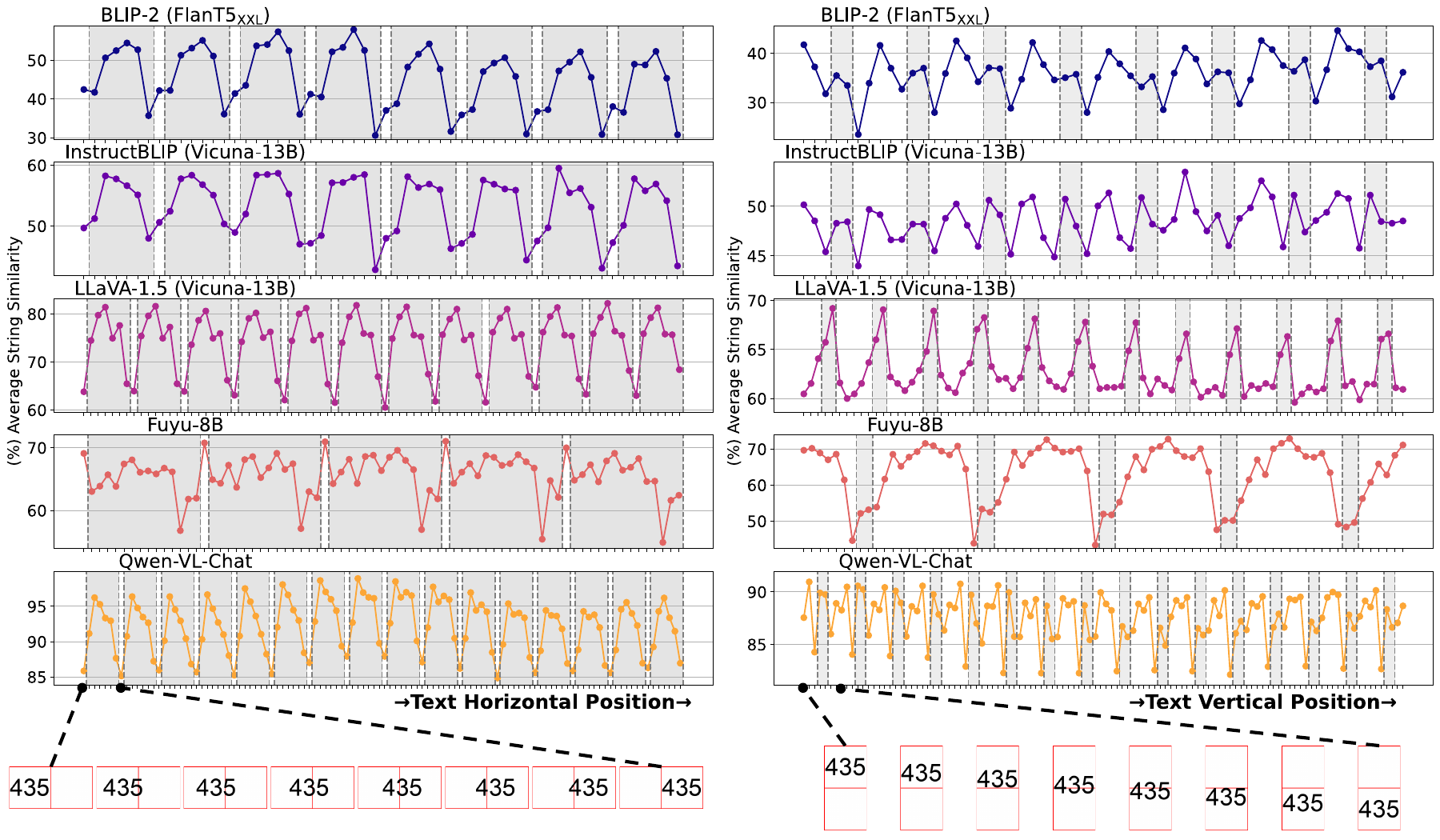}
  \caption{
  The performance of MLLMs in text recognition tasks demonstrates notable variability when textual content is vertically (left) and horizontally (right) cut by image patch boundaries. \textcolor{gray}{Gray} area indicates that the target texts are cut by a patch boundary. We provide two local illustrations below showing that a text is shifted between two adjacent image patches. Due to space constraints, we only present the middle part of the entire shifting (range ratio from 0.25 to 0.75), the complete plots are presented in the~\autoref{fig:compelete_cut}.
  }
  \label{fig:cut}
  \vspace{-0.1cm}
\end{figure*}
Small objects in an image, in addition to the inherent effect of their size we observed in the previous section, can also affect MLLMs' perception by allowing for the presence of more distractors in the image. Our goal in this section is to study the effect of distractors on MLLMs' perception of small objects. To that end, we place the answer text (number) at the center of the image, then we introduce $k$ distractor numbers, positioning them at random locations throughout the image. The answer digit text is assigned to the variable `a', while the distractor numbers are assigned to `b' and subsequent letters (`c', `d'......). We vary the number of distractors from 0 to 9, and prompt MLLMs with \textit{"What is the number assigned to variable ‘a’ in the image?"}. We experiment with text font sizes 8 and 12 without resampling to gain 2 tiers of task difficulty, each tier including 100 random numbers (3 digits) to read, and the random position of distractors for each number is varied 5 times.

\textbf{Increasing the number of distractors makes perception harder for MLLMs.}
As shown in~\cref{fig:distract}, the increase in the number of distractors consistently decreases MLLMs' performance regardless of their overall performance.
Specifically, the OCR-enhanced MLLM Qwen-VL-Chat reaches a perfect score across varying distractor numbers on font size 12, while facing a 10-point performance drop during the increase of distractor numbers on font size 8.
Among the other models, Fuyu-8B, InstructBLIP, and BLIP-2 present heightened sensitivity to the additional distractors while LLaVA keeps a relatively minor performance drop. It is worth noting that although Fuyu-8B has superior performance over LLaVA-1.5 in~\cref{fig:size}, it appears to lack robustness when facing more complex visual questions.

\vspace{-0.2cm}
\subsection{Location Sensitivity Study}
\label{sec:location}
Another factor that can significantly vary for small objects is their location in the image, which can in turn affect MLLMs' perception. We study two complementary location-related factors in this section: the global location on the image and the local patch boundary cut on the target object (described in detail at the start of~\cref{sec:problem_formulation}).

\vspace{-0.2cm}

\subsubsection{Global location}
\autoref{tab:patchsize} outlines the patch sizes and counts of the MLLMs evaluated in our study. To augment patch capacities, we amalgamate every four adjacent $14\times 14$ image patches from models like BLIP-2, InstructBLIP, LLaVA-1.5, and Qwen-VL-Chat into a single $28\times 28$ patch. Texts are centrally placed within each merged patch, maintaining a consistent sampling rate of 8.
In this experiment, following the setting of~\cref{sec:distractor}, we examine MLLMs' text recognition and localization performance under variations in distractor presence and global text positioning. For assessing MLLMs' capabilities, we introduce scenarios with zero and $k$ distractors---zero distractors that evaluate pure text recognition ability across different image locations and $k$ distractors that require localizing the target text. Specifically, the OCR-enhanced Qwen-VL-Chat model is tested with nine distractors, while all other models with one distractor. We include 100 random numbers (3 digits) placed all through the image patches. We prompt MLLMs with the question \textit{"What is the number assigned to variable ‘a’ in the image?"} during evaluation.

\textbf{MLLMs exhibit inconsistent text recognition and localization performance across different global locations.} 
It is observed that the majority of models, except LLava-1.5, encounter challenges in recognizing or localizing text on the right side of an image. Moreover, BLIP2 and InstructBLIP also experience difficulties with text on the left side. Notably, the OCR-enhanced model Qwen-VL-Chat, despite obtaining a near-perfect score in most locations, demonstrates a significant performance disparity of 58 points across different locations. Also, Fuyu-8B experiences a sharp decrease in its performance in the first row.
This observation suggests that MLLMs are susceptible to positional bias when processing images. While including more training datasets can lead to much better overall performance, performance drops on certain image regions still exist.

\vspace{-0.2cm}
\subsubsection{Local Patch Boundary Cut}
We construct a dataset where the generated digital text gradually crosses an image boundary. For vertical patch boundary cut, the digit text is anchored at a predetermined vertical location, while being horizontally moved across the full span of the image. For horizontal cuts, the digit text is fixed at a specific horizontal position and moved vertically. An illustrative example of vertical cut is shown at the bottom of~\autoref{fig:cut}.
We determine the number of reading digits depending on the maximum digit capacity for a single image patch, specifically setting at six digits for Fuyu-8B and three digits for the remaining models. We include 100 random numbers for each experiment. We prompt MLLMs with \textit{"What is the number on the image?"} during evaluation.

\textbf{Model's performance is lower when target objects remain undivided by patch boundaries.}
For image patch boundary vertical cutting,
as observed in~\autoref{fig:cut} (left), a common trend among all models is the performance decline at the center of the patch, where texts remain undivided by patch boundaries(white parts). Notably, although presenting a near-perfect score, Qwen-VL-Chat still presents an around 10 percent gap between different patch boundary cuts. The only model that does not show this trend is Fuyu-8B - we assume this is due to its enlarged patch size, making the performance inside an image patch more robust.
This phenomenon indicates that contrary to intuitions, texts divided across multiple patches may be more effectively recognized by MLLMs. Therefore, even with the same size and quality, small objects seem to be more recognizable by MLLMs when they are divided into different image patches.

\textbf{Horizontal cuts hurt the performance more than vertical cuts.}
\autoref{fig:cut} (right) demonstrates the performance of the five models when the target text is horizontally cut by a patch boundary. Consistent with vertical cuts, in LLaVA-1.5, we observe a notable performance peak at the boundary cuts. However, the remaining models do not show such a trend. We hypothesize two factors contributing to this observation. First, at the horizontal cut, all characters presented are divided into two separate parts, while the vertical cut divides at most only one character into different patches. This effect potentially diminishes the completeness of shape information. Second, for the horizontal cut, the two resulting image tokens are positioned further apart after the image is translated into sequence input of transformers; for the vertical cut, the two corresponding image patches remain continuous in the resulting sequence.

\vspace{-0.2cm}
\section{Conclusion}

In this paper, we expose notable limitations of current MLLMs on perceiving small visual details. To gain a further understanding of the limitation, we identify four independent relevant factors: object quality, size, distractors, and location. We extensively explore the effect of each of the factors by conducting carefully controlled intervention studies. Based on our study, we suggest that: 1) Object quality does not pose an additional obstacle for MLLMs after a certain threshold, however, object quality should be carefully considered when images are resampled before feeding to MLLMs; 2) most MLLMs fall short in perceiving small objects, even with enough object quality, explicit training could potentially overcome this gap; 3) MLLMs' performance is significantly affected by the target objects' global location in the image; 4) MLLMs are good at recognizing small objects that are divided into more image tokens, while horizontal cutting could hurt performance due to the distance between vertically adjacent image patches. In addition to the findings, our study also provides a new evaluation protocol for the future enhancement of MLLMs' perception.

\clearpage
\section*{Impact Statement}
This research identifies and analyzes critical limitations in Multimodal Large Language Models (MLLMs) regarding the recognition of small visual details, emphasizing the roles of object quality, size, distractors, and location. Our findings will potentially offer insights to improve visual processing capabilities. The introduction of a new evaluation protocol provides a foundation for future advancements, aiming to enhance MLLMs' applicability in diverse real-world scenarios. This work contributes to the development of more robust and reliable MLLMs.




\bibliography{ref}
\bibliographystyle{icml2024}

\newpage
\appendix
\onecolumn

\section{Complete result of patch boundary cut.}
\begin{figure*}[t]
  \centering
  \includegraphics[trim=0 0 0 0, clip, width=0.85\textwidth]{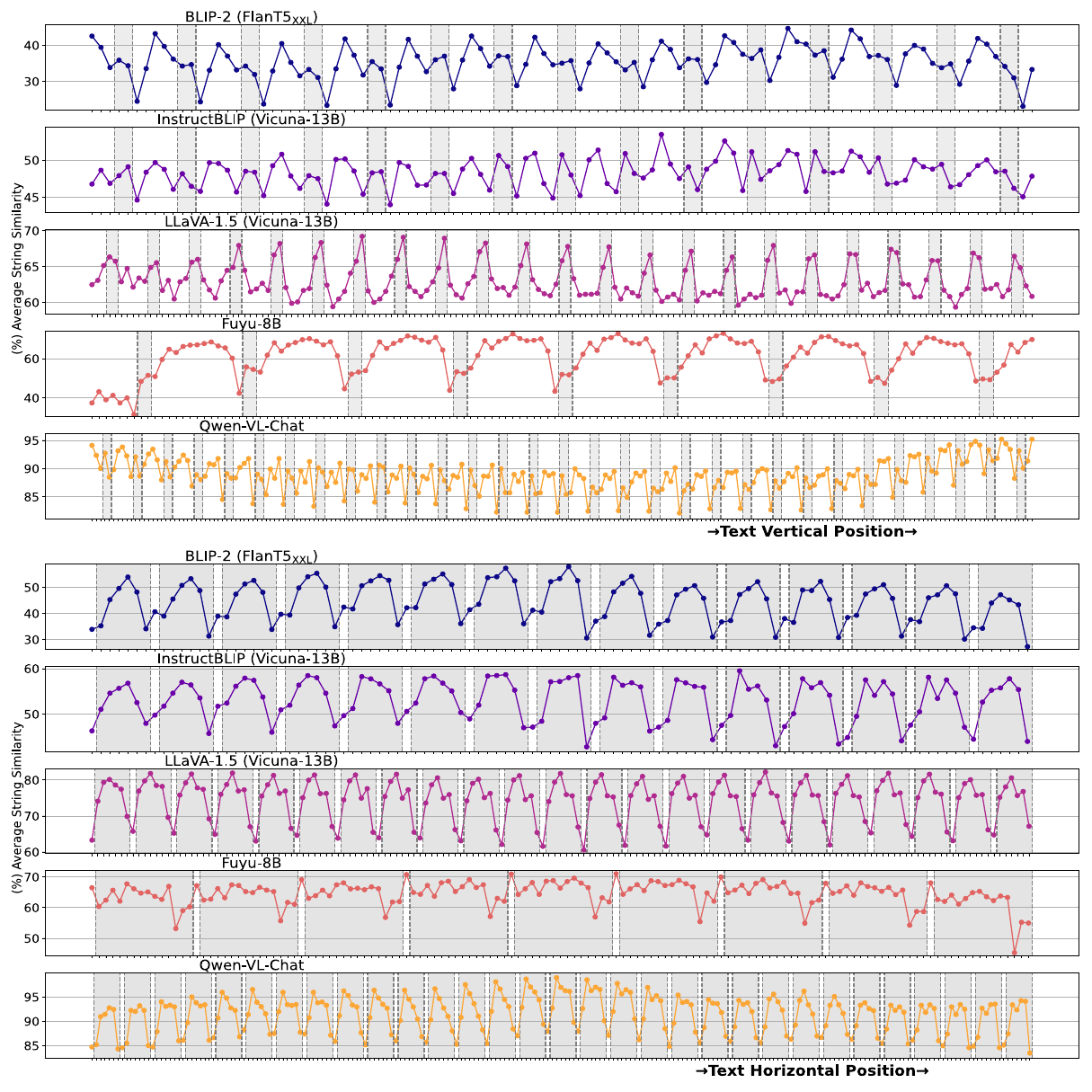}
  \caption{
  Complete result of vertical and horizontal cut.
  }
  \label{fig:compelete_cut}
  \vspace{-0.1cm}
\end{figure*}

~\autoref{fig:compelete_cut} shows the complete result of horizontal (upper) and vertical (lower) cuts, the overall trend stays the same.

\section{Result from GQA and TextVQA on different matching strategies.}
\label{sec:appendix_motivation}
\begin{figure}[t]
  \centering
  \includegraphics[trim=0 0 0 0, clip, width=0.45\textwidth]{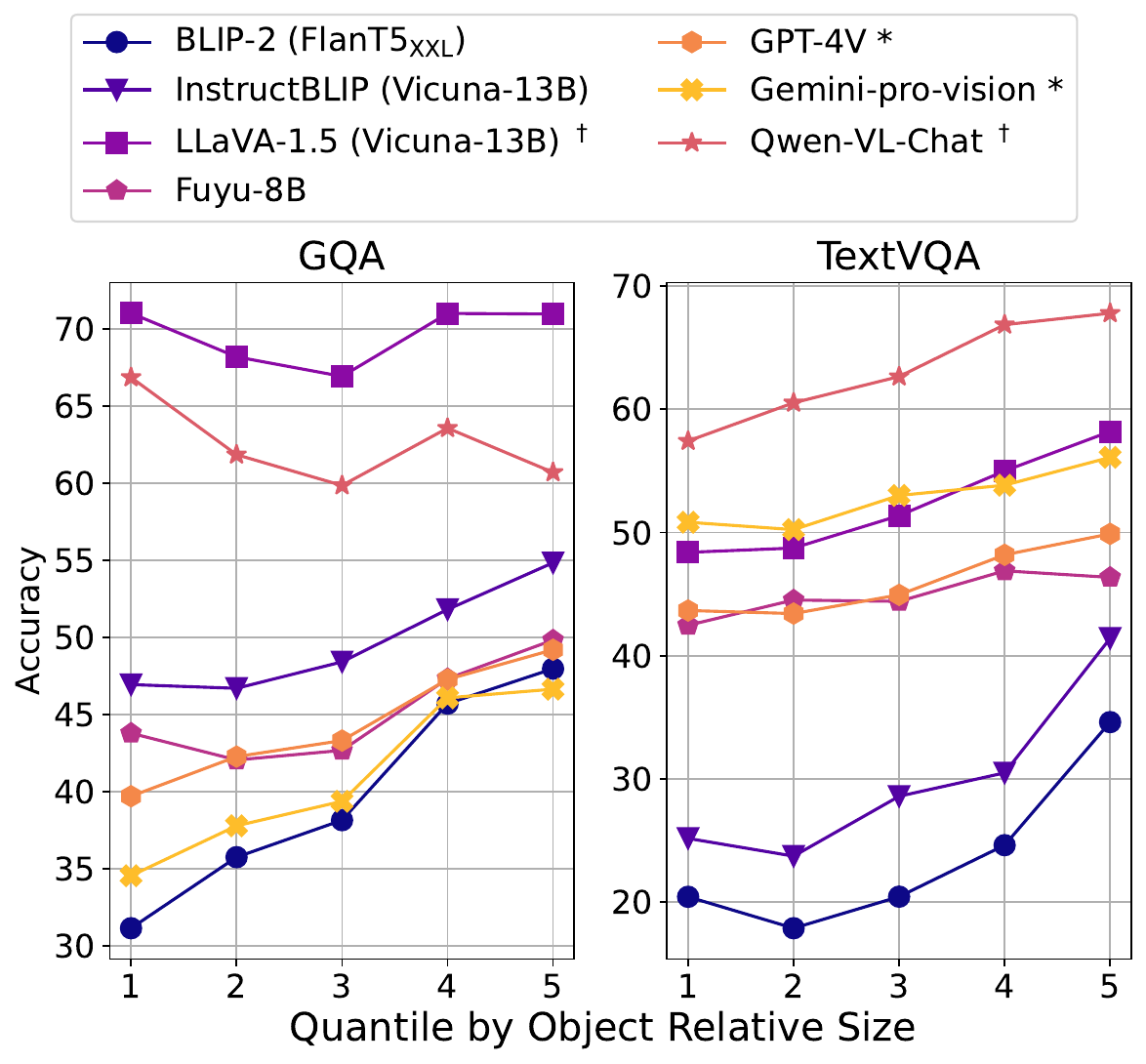}
  \caption{
  The performances of 7 MLLMs on GQA and TextVQA quantiles by object relative size. The models' predictions are computed with \textbf{exact matching}. The overall trend stays the same despite some variances in performance.
  *A small part of the dataset is skipped due to safety policy of API models. $^\dagger$The model has been reported to be trained on the dataset.
  }
  \label{fig:motivation_exact}
  \vspace{-0.1cm}
\end{figure}

In addition to inclusion matching, in~\autoref{fig:motivation_exact}, we use exact string matching to compute the accuracy. The most notable difference is that some models' performance diminishes, as their output do not follow the dataset format strictly. Despite the above, the overall trend that most of the models have difficulty perceiving smaller details stays the same.

\section{Further analysis on the role of object distractors.}
\begin{figure}[t]
  \centering
  \begin{minipage}{0.45\textwidth}
    \includegraphics[width=\linewidth]{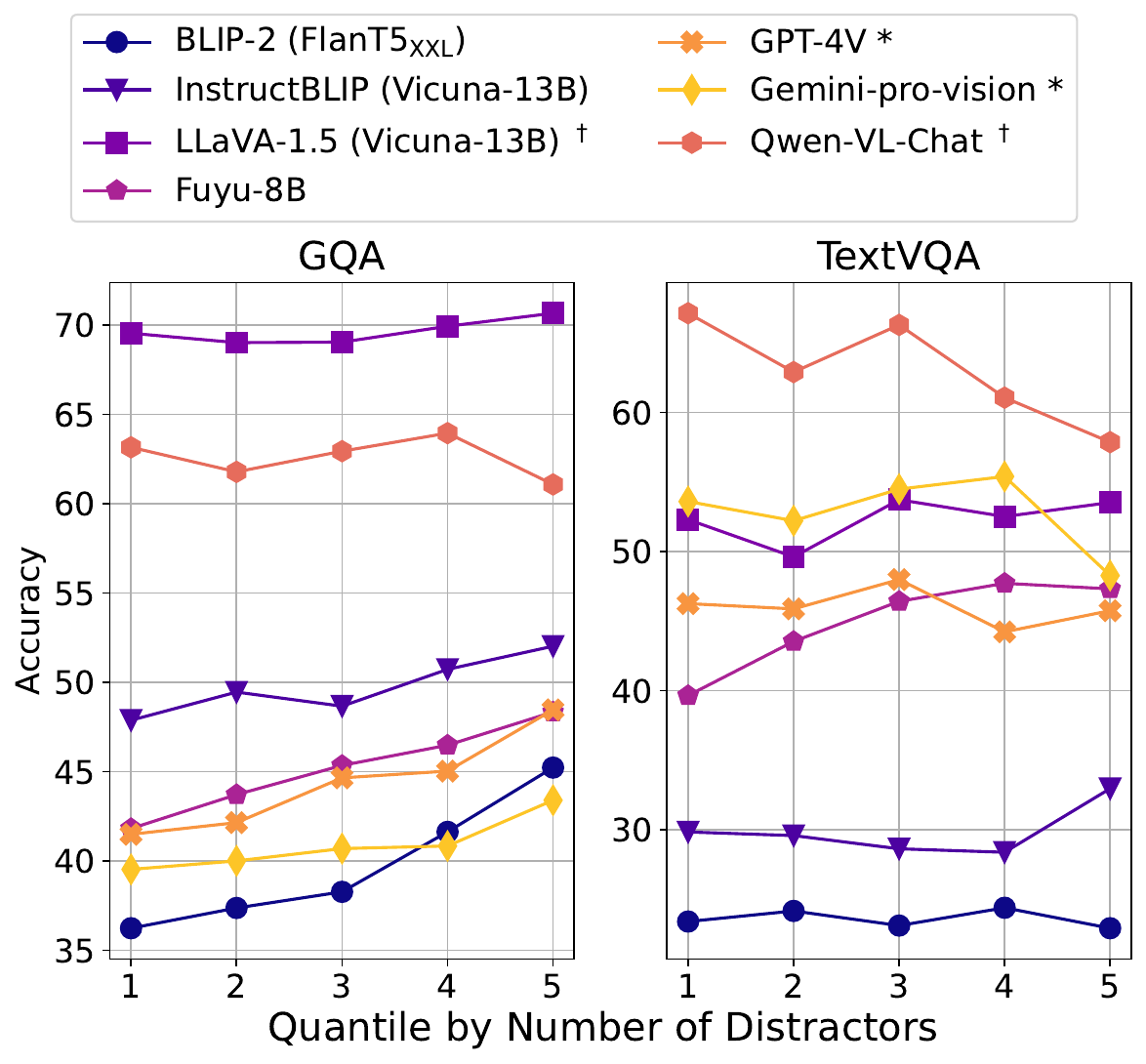}
    \caption{MLLMs' performance on TextVQA and GQA quantiles divided by number of distractors, accuracy is computed using exact matching.}
  \end{minipage}
  \hfill 
  \begin{minipage}{0.45\textwidth}
    \includegraphics[width=\linewidth]{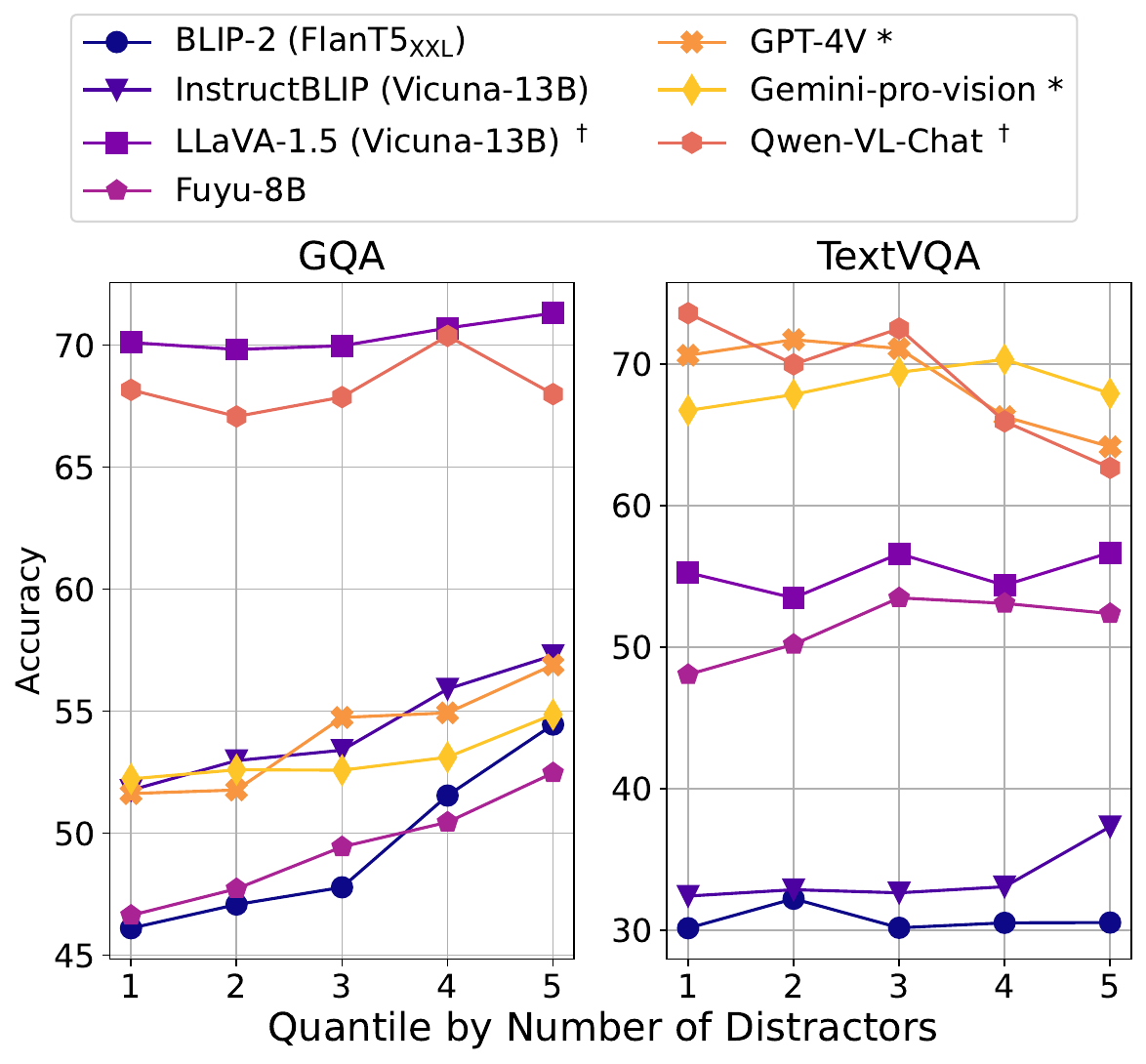}
    \caption{MLLMs' performance on TextVQA and GQA quantiles divided by number of distractors, accuracy is computed using inclusion matching.}
  \end{minipage}
  \label{fig:distract_motivation}
\end{figure}

~\autoref{tab:ranges} presents the number of distractors in each group of TextVQA and GQA, from which it's also clear that the number of distractors decreases when the object's relative size gets larger. Hence, it is unclear which factor plays the most important role in~\autoref{fig:motivation}. To this end, we divide the GQA and TextVQA into five quantiles by the number of object/OCR token distractors, the result of both metrics is presented in~\autoref{fig:distract_motivation}. 
From the plot, we observe that object distractors in GQA seem to affect the MLLM's performance, while in TextVQA, we do not observe a clear correlation between number of distractors and performance.

\section{What is causing the variance in positional bias?}
The different positional biases observed from~\cref{sec:location} may stem from the bias from textual training data. Typically, textual content in training datasets is oriented from left to right and concentrated towards the center of images. This common formatting convention may inadvertently lead to the under-representation of text located on the right side of images and along their margins. 

\section{Why does Fuyu-8B have a noticeable low performance in its first row?}
\label{sec:fuyu}
In~\autoref{fig:position}, we notice a sharp decrease in Fuyu-8B's performance score within the first row. We assume this unexpected phenomenon is related to its unique pure transformer decoder architecture. To this end, we choose several images and present the attention map of Fuyu-8B, providing observations for further investigation.
\begin{figure*}[t]
  \centering
  \includegraphics[trim=0 0 0 0, clip, width=0.95\textwidth]{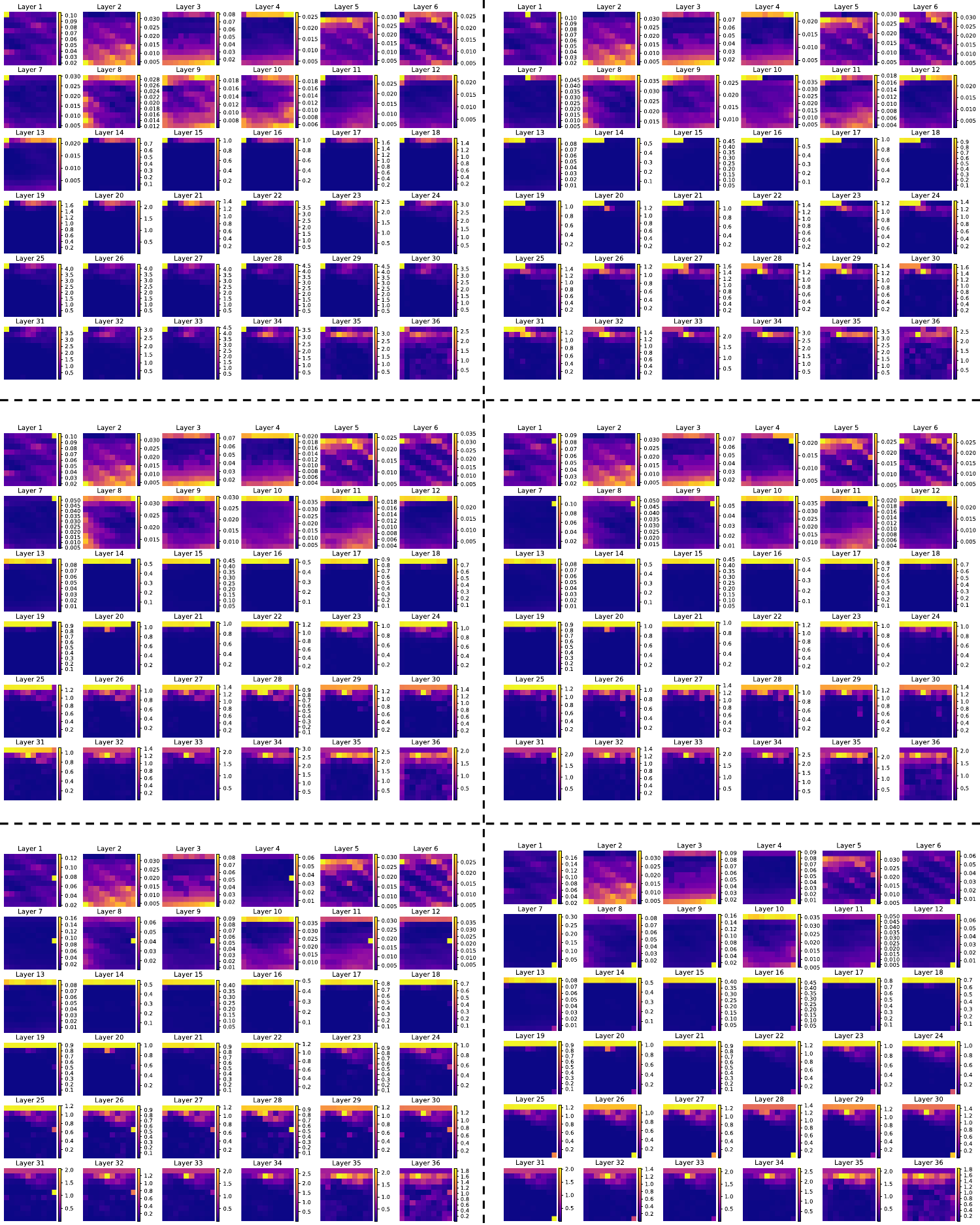}
  \caption{The attention map of Fuyu-8B on six different input images. Detailed descriptions in~\autoref{sec:fuyu}.
  }
  \label{fig:att}
  \vspace{-0.1cm}
\end{figure*}

In~\autoref{fig:att}, we provide the attention map for each of the 36 layers of Fuyu-8B. The input image is the synthetic image we construct in the location study in~\cref{sec:location}, where a single `a=665' is placed in an image patch's center. The position of the patch is: 0, 4, 9, 19, 49, 99, in the raster scan order of the original image (with $10\times10$ image tokens), the input position can also be seen from the yellow attention outlier in Layer 1. The attention map is computed for the next token after promoting Fuyu-8B with \textit{`Question: What is the number in the image? Short answer:'} and we track the attention of the next token with respect to each image patch.
From the attention map, we can tell that ranging from approximately 13-27 layers, for the image whose text is placed in $i_{th}$ position, there are consistently high attention values in the first $k$ tokens, where $k=i$ if $i\leq9$ otherwise $k=9$. Such a result could be linked to the low performance observed in the first row since the high attention among those layers stays consistent within the tokens in the first row.
For the deeper reason behind the phenomenon, we leave them as open future works.

\end{document}